\documentclass[letterpaper]{article} 
\usepackage{aaai2026}  
\usepackage{times}  
\usepackage{helvet}  
\usepackage{courier}  
\usepackage[hyphens]{url}  
\usepackage{graphicx} 
\usepackage{natbib}  
\usepackage{caption} 
\usepackage{algorithm}
\usepackage{algorithmic}
\usepackage{amsmath}
\usepackage{amsfonts}
\usepackage{booktabs}
\usepackage{multirow}
\usepackage{colortbl} 
\usepackage{xcolor}   
\usepackage{subcaption}
\usepackage{placeins}
\usepackage{newfloat}
\usepackage{listings}
\DeclareCaptionStyle{ruled}{labelfont=normalfont,labelsep=colon,strut=off} 
\floatstyle{ruled}
\newfloat{listing}{tb}{lst}{}
\floatname{listing}{Listing}
\title{Multivariate Diffusion Transformer with Decoupled Attention for \\ High-Fidelity Mask-Text Collaborative Facial Generation\thanks{This work was supported by the Natural Science Foundation of China under Grant No. 62222606.}}
\author{
    Yushe Cao\textsuperscript{\rm 1},
    Dianxi Shi\textsuperscript{\rm 2} \thanks{Co-Corresponding Author.},
    Xing Fu\textsuperscript{\rm 3},
    Xuechao Zou\textsuperscript{\rm 4},\\
    Haikuo Peng\textsuperscript{\rm 5},
    Xueqi Li\textsuperscript{\rm 2},
    Chun Yu\textsuperscript{\rm 1}
    Junliang Xing\textsuperscript{\rm 1}\thanks{Co-Corresponding Author.},
}
\affiliations{
    \textsuperscript{\rm 1}Tsinghua University \\
    \textsuperscript{\rm 2}Intelligent Game and Decision Lab\\
    \textsuperscript{\rm 3}Independent Researcher \\
    \textsuperscript{\rm 4}Beijing Jiaotong University \\
    \textsuperscript{\rm 5}National University of Defense Technology\\
    cao-ys23@mails.tsinghua.edu.cn, xuechaozou@foxmail.com, jlxing@tsinghua.edu.cn  
}

\usepackage{bibentry}

\begin{document}

\maketitle

\begin{abstract}
While significant progress has been achieved in multimodal facial generation using semantic masks and textual descriptions, conventional feature fusion approaches often fail to enable effective cross-modal interactions, thereby leading to suboptimal generation outcomes. To address this challenge, we introduce MDiTFace—a customized diffusion transformer framework that employs a unified tokenization strategy to process semantic mask and text inputs, eliminating discrepancies between heterogeneous modality representations. The framework facilitates comprehensive multimodal feature interaction through stacked, newly designed multivariate transformer blocks that process all conditions synchronously. Additionally, we design a novel decoupled attention mechanism by dissociating implicit dependencies between mask tokens and temporal embeddings. This mechanism segregates internal computations into dynamic and static pathways, enabling caching and reuse of features computed in static pathways after initial calculation, thereby  reducing additional computational overhead introduced by mask condition by over 94\% while maintaining performance. Extensive experiments demonstrate that MDiTFace significantly outperforms other competing methods in terms of both facial fidelity and conditional consistency.
\end{abstract}


\section{Introduction}
\label{sec:introduction}
High-fidelity facial synthesis \cite{ning2023multi,chen2024dreamidentity,melnik2024face,wang2025facea} holds transformative potential across digital identity systems, adversarial security, and immersive media applications \cite{terhorst2023qmagface,rehaan2024face,diao2025ft2tf}. While recent advances in generative modeling, such as GANs and diffusion models, have enabled photorealistic facial generation, deploying these technologies in practice remains hindered by in fine-grained controllability, posing challenges in precisely generating facial images that align with user expectations.

Modern conditional generation frameworks have transitioned from unimodal to multimodal paradigms to tackle these challenges. Initially, text-driven methods constructed end-to-end pipelines to transform semantic descriptions into facial attributes \cite{kim2022diffusionclip,zhu_grainedclip_2023}. However, they encountered difficulties in achieving precise spatial localization (e.g., pose, geometry) due to the inherent ambiguity of language. Subsequently, visual modalities such as semantic masks \cite{chen2022sofgan,ergasti2024controllable} were incorporated to enforce geometric constraints, but this often came at the cost of reduced semantic versatility in attributes like age, gender, or complexion. This dichotomy has spurred research into multimodal collaborative facial synergies \cite{meng2025mm2latent,sowmya2024generative,kim2024diffusion} that leverage the expressive capacity of natural language for high-level attributes and the precision of visual cues for low-level spatial control.

Existing state-of-the-art multimodal facial synthesis methods generally adhere to two primary architectural paradigms: (1) Generative Adversarial Network (GAN)-based frameworks \cite{du2023pixelface+,kim2024diffusion,meng2025mm2latent}, such as StyleGAN variants, which project heterogeneous modality conditions into a latent space via multiple encoding networks for subsequent fusion; and (2) diffusion models \cite{nair2023unite,huang2023collaborative,peng2024controlnext,kim2024diffusion}, which integrate unimodal expert diffusion models through dynamic weighting during the inference process. However, these approaches typically process multiple modalities in isolation or employ simplistic feature fusion strategies, resulting in suboptimal multimodal feature interaction—particularly in scenarios that demand collaborative optimization between visual masks and textual descriptions.

In this work, we propose a customized diffusion transformer framework termed MDiTFace, which aims to tackle the challenge of inadequate cross-modal feature interaction in high-fidelity mask-text collaborative facial synthesis. The framework employs a unified tokenization approach to process multimodal conditions from mask and text inputs and projects them into a shared latent space via modality-specific embedders, effectively bridging the gap between heterogeneous modality representations. We design a dedicated multivariate transformer block capable of simultaneously processing multi-conditional token sequences derived from both text and mask modalities. The internal attention computation within this block significantly enhances cross-modal feature interaction capabilities. Additionally, by decoupling the implicit dependency between mask tokens and temporal embeddings, we introduce a novel decoupled attention mechanism. It explicitly splits the internal computational flow into dynamic and static pathways, enabling relevant features in the static pathway to be cached and reused across denoising steps after initial computation. This approach reduces the additional computational overhead introduced by mask condition by over 94\% while maintaining model performance.

In summary, our core innovations and contributions are as follows:
\begin{itemize}
    \item We propose a customized diffusion transformer framework, MDiTFace, which resolves the issue of insufficient multimodal feature interaction in high-fidelity mask-text collaborative facial synthesis, significantly improving the quality of synthesized facial images.
    \item We design a dedicated multivariate transformer block that can simultaneously process multimodal token sequences from both text and mask inputs, with its internal attention computation substantially facilitating bidirectional and flexible interaction between multimodal features.
    \item We introduce a novel decoupled attention mechanism that explicitly divides the internal computational flow into dynamic and static pathways. By reusing computational features in the static pathway, it reduces the computational overhead associated with mask tokens by over 94\% while preserving model performance..
\end{itemize}
Comprehensive qualitative and quantitative evaluations demonstrate that MDiTFace achieves state-of-the-art performance in both facial fidelity and multimodal conditional consistency.

\begin{figure}[t]
\centering
\includegraphics[width=0.9\columnwidth]{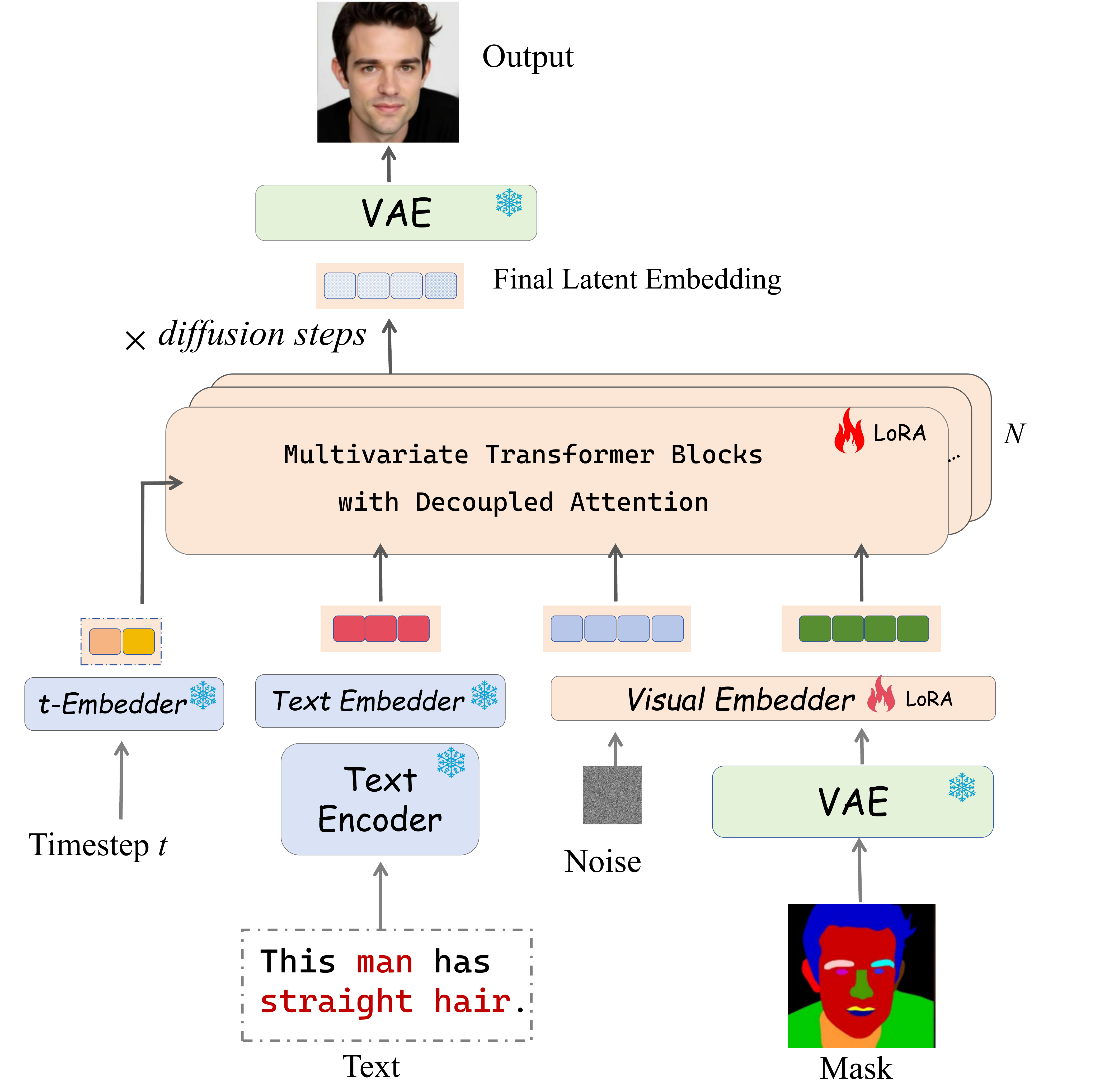} 
\caption{Overall framework of our MDiTFace method.}
\label{fig:framework}
\end{figure}

\section{Related Works}
\subsection{Diffusion Models for Image Generation}
Recent advancements in generative modeling have witnessed a paradigm shift from GAN-based architectures \cite{you2022application,du2023pixelface+,che2024assessing,cao2026dual} to diffusion models \cite{po2024state,he2025diffusion,chang2025design,huang2025diffusion}, driven by their superior training stability and output diversity.  Early diffusion models like DDPM \cite{ho2020denoising} and DDIM \cite{song2020denoising} established foundational sampling pipelines, while subsequent works focused on computational efficiency \cite{rombach2022high,xue2024accelerating,chadebec2025flash} and conditional control \cite{podell2023sdxl,yoon2023score,peng2024controlnext}. Latent diffusion models (LDM) \cite{rombach2022high} reduced computational costs by operating in compressed latent spaces, enabling high-resolution synthesis. ControlNet \cite{zhang2023adding}, ControlNeXt \cite{peng2024controlnext}, and T2I-Adapter \cite{mou2024t2i} further enhanced controllability by introducing external conditioning networks, though their multimodal integration is still confined to the addition of independent feature maps.

The integration of vision transformers has driven recent breakthroughs, where diffusion transformers (DiTs) \cite{peebles2023scalable,wang2025lavin} demonstrate superior performance over U-Net architectures \cite{ronneberger2015u,cao2024mefusion} through global attention mechanisms, with representative works including FLUX.1 \cite{flux2024}, OminiControl \cite{tan2024ominicontrol,tan2025ominicontrol2}, and OmniGen \cite{xiao2025omnigen}. However, these methods often incur significantly higher computational costs.
\subsection{Facial Image Synthesis with Multimodal Conditioning}
Facial synthesis techniques \cite{kim2023dcface,melnik2024face,song2025attridiffuser} have evolved from unimodal control paradigms to multimodal control paradigms, with recent research efforts focusing on the collaborative integration of various modalities, such as text and semantic masks \cite{du2023pixelface+,meng2025mm2latent}. Methods like TediGAN \cite{xia2021tedigan} and MM2Latent \cite{meng2025mm2latent} leverage the disentangled properties of the $\mathcal{W}$ latent space in StyleGANs to achieve multimodal-driven facial synthesis by fusing latent vectors from multiple modalities. In contrast, diffusion-based approaches, such as GCDP \cite{park2023learning}, have aimed to enhance semantic alignment in synthesized faces; however, they are limited by the restrictive assumptions of joint Gaussian distributions. Additionally, methods like UaC \cite{nair2023unite} and CoDiffusion \cite{huang2023collaborative} dynamically integrate multiple unimodal expert models during inference, which results in a more than two-fold increase in computational cost. Meanwhile, DDG \cite{kim2024diffusion} combines diffusion models with GANs, utilizing ControlNet \cite{zhang2023adding,peng2024controlnext} for semantic mask and text multimodal feature fusion. Still, the expressiveness of the StyleGAN latent space representation constrains the quality of synthesized faces.

\section{Method}
\label{sec:methods}

\begin{figure*}[t]
\centering
    \centering
    \includegraphics[width=\linewidth]{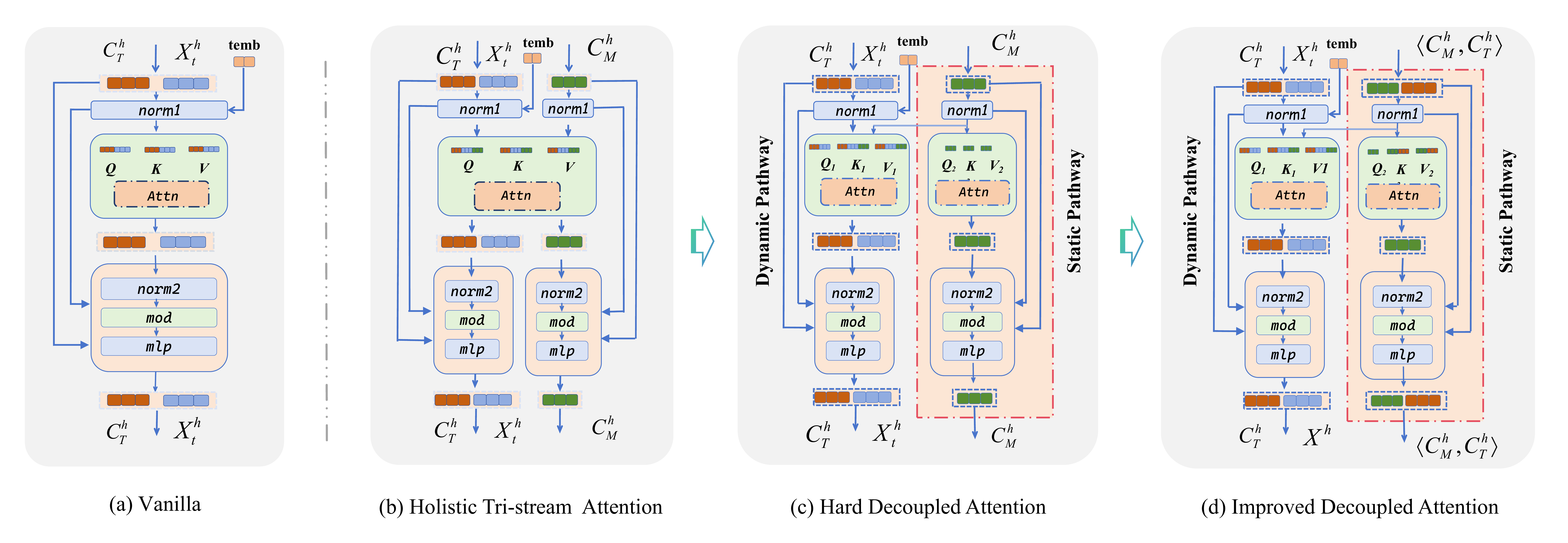} 
    \caption{Internal attention design of the multivariate transformer block. (a) The vanilla dual-stream attention in FLUX.1, which exclusively supports text-modal conditioning. (b) Extended holistic tri-stream attention supporting mask-text multimodal conditions at significantly increased computational cost; (c) Hard-decoupled attention with dynamic and static pathways,efficiency improves, but at the cost of performance degradation; (d) Improved decoupled attention restoring mask-to-text perceptual pathways for balanced efficiency and model performance.}
    \label{fig:mm-attention-evolutions}
\end{figure*}

\subsection{Preliminaries}
Diffusion transformer (DiT) \cite{peebles2023scalable} successfully breaks through the limitations of local perception by adopting a pure transformer architecture to replace the traditional U-Net backbone network. This achieves an integration of global dependency modeling and multimodal information fusion, thereby significantly enhancing the model's capability to model semantic consistency in complex scenarios. At time step $t$, the model utilizes stacked tansformer blocks to synchronously process noisy image tokens $\mathbf{X}_t \in \mathbb{R}^{N \times d}$ and text condition tokens $\mathbf{C}_T \in \mathbb{R}^{L \times d}$. 
Each transformer block incorporates a multi-head attention computation via Equation \ref{eq:attention}, ensuring in-depth interaction and fusion between visual and semantic information.
\begin{equation}
\text{Attn}(\mathbf{Q}, \mathbf{K}, \mathbf{V}) = \text{Softmax}\left(\frac{\mathbf{Q}\mathbf{K}^\top}{\sqrt{d_h}}\right)\mathbf{V},
\label{eq:attention}
\end{equation}
where $\mathbf{Q}$, $\mathbf{K}$, and $\mathbf{V}$ are linearly projected from the concatenated tokens $[\mathbf{C}_T; \mathbf{X}_t]$, and $d_h$ denotes the head dimension.
 
\textbf{Rotary Positional Embedding.}
To overcome the translation invariance limitation of traditional absolute position encoding, DiT introduces Rotary Position Encoding (RoPE) \cite{heo2024rotary} to explicitly model spatial relationships. Each visual token at position $(i,j)$ is modulated via:
\begin{equation}
    \mathbf{X}_{i,j} \leftarrow \mathbf{X}_{i,j} \cdot \mathbf{R}(i,j),
    \label{eq:rope_modulation}
\end{equation}
where $\mathbf{R}(i,j)$ denotes frequency-domain rotation. Text tokens are assigned (0,0) to maintain semantic coherence.
\subsection{Model Design}
Figure \ref{fig:framework} illustrates the overall framework of the proposed MDiTFace method, which aims to synthesize high-fidelity facial images under the collaborative constraints of textual descriptions and semantic masks. Built upon the FLUX.1 text-to-image generation model \cite{flux2024,labs2025flux1kontextflowmatching}, MDiTFace employs a unified tokenization strategy to process heterogeneous mask-text conditional inputs, projecting them through modality-specific embedders into a shared latent space where the noise image tokens $\mathbf{X}_t$ resides, thereby eliminating representation disparities across heterogeneous modalities. To achieve synchronous processing and achieve a thorough fusion of multiple conditional token sequences, which include mask tokens $C_M$ and text tokens $C_T$, we have conducted a comprehensive redesign and subsequent implementation of the multivariate transformer block. These blocks incorporate an innovatively designed decoupled attention mechanism that enables flexible interaction among multimodal tokens while partitioning internal computations into dynamic and static pathways. By eliminating the implicit dependency between mask tokens $C_M$ and temporal embeddings, the relevant features of the static pathway can be cached after the initial computation and reused across denoising steps, thereby significantly reducing redundant computational overhead.

\textbf{Unified Tokenization.}
To achieve collaborative facial synthesis combining semantic masks and textual descriptions, the key challenge lies in efficiently integrating semantic mask conditions into the foundational model. An intuitive approach is to directly concatenate the VAE-encoded mask condition $\mathbf{C}_M$ with the noisy image $\mathbf{X}_t$ along the channel dimension, as shown in Equation \ref{eq:mask_concat}:
\begin{equation}
\mathbf{X}_t \leftarrow \text{Concat}(\mathbf{X}_t, \mathbf{C}_M).
\label{eq:mask_concat}
\end{equation}
However, experimental results indicate that this simple fusion strategy leads to poor mask-condition consistency in synthesized facial images (see Table \ref{tab:mask_condition_processing} for details). To address this issue, our proposed MDiTFace method employs a unified tokenization strategy to process multimodal input conditions: The mask is encoded by reusing the VAE encoder, preserving spatial position information while maintaining architectural simplicity. The text is encoded using the pre-trained T5 encoder \cite{raffel2020exploring} to capture semantic richness. Subsequently, the encoded features from both modalities are projected into a latent space shared with the noisy image tokens $\mathbf{X}_t$ via modality-specific embedders, bridging representation gaps between heterogeneous modalities. This transformation is formalized in Equation \ref{eq:unified_tokenization}:
\begin{equation}
\begin{aligned}
\mathbf{C}_M &= \text{VisualEmbedder}(\mathbf{VAE}(\text{Mask})) \in \mathbb{R}^{N \times d}, \\
\mathbf{C}_T &= \text{TextEmbedder}(\mathbf{T5}(\text{Text})) \in \mathbb{R}^{L \times d}.
\end{aligned}
\label{eq:unified_tokenization}
\end{equation}
To strengthen spatial alignment, we apply the same RoPE to the mask tokens $\mathbf{C}_M$ as to the noisy image tokens $\mathbf{X}_t$, ensuring precise correspondence between them in the spatial dimension. The operation is defined in Equation \ref{eq:mask_rotation}:
\begin{equation}
\mathbf{C}_{M_{i,j}} \leftarrow \mathbf{C}_{M_{i,j}} \cdot \mathbf{R}(i,j).
\label{eq:mask_rotation}
\end{equation}
Finally, a joint token sequence $[\mathbf{C}_T; \mathbf{X}_t; \mathbf{C}_M]$ is constructed by concatenating text tokens $\mathbf{C}_T$, noisy image tokens $\mathbf{X}_t$, and mask tokens $\mathbf{C}_M$ along the sequence dimension. This unified tokenization provides a foundational representation for further interaction of multimodal features.

\textbf{Multivariate Transformer Block.}
In the vanilla transformer block, its inherent dual-stream token processing paradigm is primarily designed for text-driven image synthesis tasks (as shown in Figure~\ref{fig:mm-attention-evolutions})(a), making it difficult to directly accommodate the synchronous processing requirements of multimodal conditional tokens (mask tokens $\mathbf{C}_M$ and text tokens $\mathbf{C}_T$). To address this limitation, we have specifically reconstructed the internal architecture of the transformer block, upgrading its intrinsic dual-stream attention mechanism to a tri-stream attention architecture (as depicted in Figure~\ref{fig:mm-attention-evolutions}(b)) to support the collaborative processing of multiple conditional tokens.

Specifically, this module performs linear projections on the inputs of the three modalities separately using independent parameter matrices to generate queries ($\mathbf{Q}$), keys ($\mathbf{K}$), and values ($\mathbf{V}$). The calculation process is as follows:
\begin{equation}
\begin{aligned}
\mathbf{Q} &= [\mathbf{W}_q^T \mathbf{C}_T; \mathbf{W}_q^X \mathbf{X}_t; \mathbf{W}_q^M \mathbf{C}_M], \\
\mathbf{K} &= [\mathbf{W}_k^T \mathbf{C}_T; \mathbf{W}_k^X \mathbf{X}_t; \mathbf{W}_k^M \mathbf{C}_M], \\
\mathbf{V} &= [\mathbf{W}_v^T \mathbf{C}_T; \mathbf{W}_v^X \mathbf{X}_t; \mathbf{W}_v^M \mathbf{C}_M].
\end{aligned}
\label{eq:multivariate-transformer-blocks}
\end{equation}
Through tri-stream attention computation, noisy image tokens $\mathbf{X}_t$, text tokens $\mathbf{C}_T$, and mask tokens $\mathbf{C}_M$ can achieve sufficient dynamic interaction and feature fusion, effectively enhancing the semantic correlation among multimodal features. However, the introduction of multiple conditional tokens inevitably increases the computational complexity during the inference phase, particularly when handling high-resolution facial synthesis tasks, significantly constraining the generation efficiency.

\textbf{Decoupled Attention Mechanism.}
To effectively mitigate the increase in computational overhead, we innovatively propose a decoupled attention mechanism. Its initial design is illustrated in Figure \ref{fig:mm-attention-evolutions}(c), where the attention computation is decoupled into a dynamic pathway and a static pathway through structural design. Specifically, the dynamic pathway incorporates the perception of mask features while retaining the ability to be modulated by timestep embedding. The static pathway focuses solely on the self-attention computation of mask tokens $\mathbf{C}_M$ and decouples its dependency on the diffusion timesteps. It only needs to be computed once at the initial stage of diffusion, and the cached features can be directly reused in subsequent denoising steps, thereby significantly reducing computational complexity. However, experiments (as shown in Table \ref{tab:effectiveness_of_decoupled_attention}) reveal that this hard-decoupled design disrupts the effective perception of text tokens $\mathbf{C}_T$ by mask tokens $\mathbf{C}_M$, leading to a significant decline in mask-condition consistency of the synthesized facial images.

To address this deficiency, we optimize and improve the structure of the decoupled attention mechanism, as depicted in Figure \ref{fig:mm-attention-evolutions}(d). We use the concatenated sequence of multimodal condition tokens $\langle\mathbf{C}_M,\mathbf{C}_T\rangle$ as the input for static pathway computation, thereby restoring the perception from mask tokens $\mathbf{C}_M$ to text tokens $\mathbf{C}_T$ and ensuring the integrity and effectiveness of feature interaction. The final computational definitions of the dynamic pathway and static pathway are as follows:

\textbf{Dynamic pathway (Modulated by timesteps)}:
\begin{equation}
\begin{aligned}
    \mathbf{Q_1} &= [\mathbf{W}_q^T \mathbf{C}_T;\mathbf{W}_q^X\mathbf{X}_t], \\
    \mathbf{K_1} &= [\mathbf{W}_k^T \mathbf{C}_T;\mathbf{W}_k^X\mathbf{X}_t;\mathbf{W}_k^T \mathbf{C}_M], \\
    \mathbf{V_1} &= [\mathbf{W}_v^T \mathbf{C}_T;\mathbf{W}_v^X\mathbf{X}_t;\mathbf{W}_v^T \mathbf{C}_M]. \\
\end{aligned}
\label{eq:decoupled_QKV-1}
\end{equation}
\begin{equation}
    \text{Attn}\left(\left[\mathbf{C}_T;\mathbf{X}_t\right]\right) = \text{Softmax}\left(\frac{\mathbf{Q}_1 \mathbf{K}_1^\top}{\sqrt{d_h}}\right) \mathbf{V}_1, 
\label{eq:attention_computation-1}
\end{equation}

\textbf{Static pathway (Cache and reuse after computation)}:
\begin{equation}
\begin{aligned}
    \mathbf{Q_2} &= [\mathbf{W}_q^T \mathbf{C}_M;\mathbf{W}_q^T \mathbf{C}_T], \\
    \mathbf{K_2} &= [\mathbf{W}_k^T \mathbf{C}_M;\mathbf{W}_k^T \mathbf{C}_T], \\
    \mathbf{V_2} &= [\mathbf{W}_v^T \mathbf{C}_M;\mathbf{W}_v^T \mathbf{C}_T]. 
\end{aligned}
\label{eq:decoupled_QKV-2}
\end{equation}

\begin{equation}
    \text{Attn}([\mathbf{C}_M;\mathbf{C}_T]) = \text{Softmax}\left(\frac{\mathbf{Q}_2 \mathbf{K}_2^\top}{\sqrt{d_h}}\right) \mathbf{V}_2.
\label{eq:attention_computation-2}
\end{equation}

\subsection{Computational Analysis}
We conduct a computational complexity analysis of the attention calculations across different architectures depicted in Figure~\ref{fig:mm-attention-evolutions}. Taking $\mathbf{X}_t, \mathbf{C}_M \in \mathbb{R}^{N \times d}$ and $\mathbf{C}_T \in \mathbb{R}^{L \times d}$ as examples, where $N$ denotes the number of image tokens and $L$ represents the number of text tokens. Suppose a total of $T$ denoising timesteps are executed during inference. The computational complexity of the architecture in Figure~\ref{fig:mm-attention-evolutions}(a) is $\mathcal{O}(T \cdot (N+L)^2)$. Due to the introduction of mask tokens $\mathbf{C}_M$, the computational complexity of the architecture in Figure~\ref{fig:mm-attention-evolutions}(b) increases significantly to $\mathcal{O}(T \cdot (2N+L)^2)$, which becomes infeasible when $N$ is large.  In contrast, our optimized decoupled attention mechanism in Figure~\ref{fig:mm-attention-evolutions}(d) benefits from the fact that computations on the static pathway need to be performed only once, reducing its computational complexity to $\mathcal{O}\big(T \cdot (N+L) \cdot (2N+L) + (N+L)^2\big)$. Given that in high-resolution facial synthesis scenarios where $N \gg L$ , this design optimization can reduce the overall computational load by nearly 50\%.

\begin{figure*}[tb]
\centering
\includegraphics[width=0.98\textwidth]{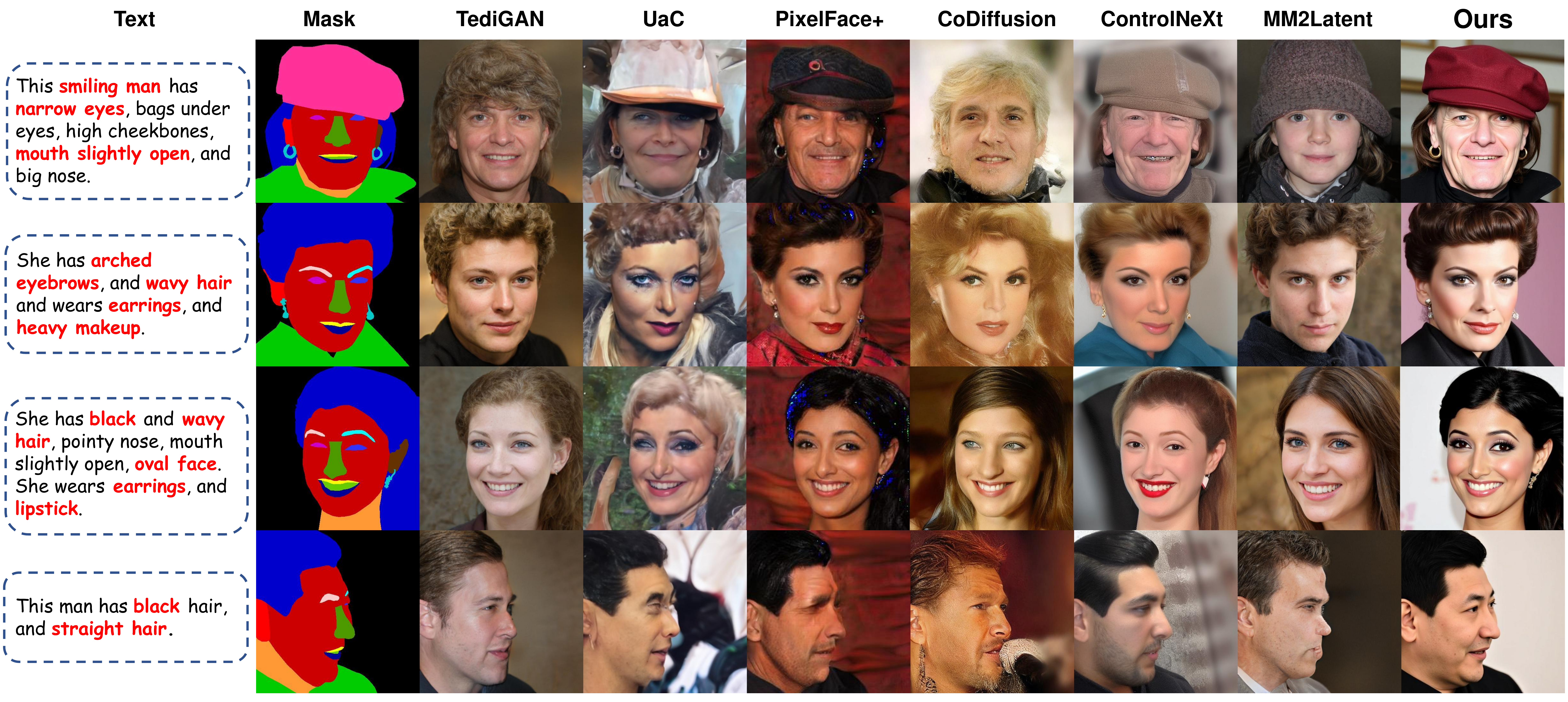} 
\caption{Qualitative comparison with state-of-the-art methods of mask-text collaborative facial generation.}
\label{fig:comparision_with_sota_methods}
\end{figure*}
 
\subsection{Training Strategy}
During the training phase, we adopt a lightweight adaptation architecture and optimize the model with the aid of flow matching loss, while introducing a stochastic condition dropout mechanism to enhance the model's performance and generalization capability.

\textbf{Training Objective.}
We employ a lightweight adaptation architecture, which only requires low-rank adaptation (LoRA) \cite{hu_lora:_2021} fine-tuning on the visual embedder and some linear layers within the multivariate transformer blocks. This approach results in training fewer than 0.1\% additional parameters, thereby significantly reducing the computational cost of training.
The model is optimized using the flow matching loss, with the specific formula as follows:
\begin{equation}
 \mathcal{L}(\theta) = \mathbb{E}_{t,\epsilon,X_t,C_T,C_M}\biggl[\| v_{\theta}(X_t,C_T,C_M,t) - \mu_t(X_t)\|_2^2\biggr],
\label{eq:flow_matching}
\end{equation}
where, $\theta$ denotes the trainable parameters, $v_{\theta}(*)$ signifies the predicted velocity field and $\mu_t(*)$ represents the target velocity field.

\textbf{Stochastic Condition Dropout.}
To further augment the generalization ability of our model, enabling it to accommodate both multi-condition collaborative facial synthesis (integrating both mask and text inputs) and single-condition facial synthesis (utilizing either text or mask input), we incorporate a stochastic condition-dropping mechanism during the training phase. Specifically, for either the text condition or the mask condition, we randomly drop them with a predefined probability $p$ (set to $p=0.1$ in our experiments). The detailed implementation can be found in Equation \ref{eq:random_drop}.
\begin{equation}
\begin{aligned}
    C_{T/M} &= 
        \begin{cases} 
            \phi & \text{with probability } p, \\
            C_{T/M} & \text{with probability } 1 - p.
        \end{cases} 
\end{aligned}
\label{eq:random_drop}
\end{equation}

\section{Experiments}
\label{sec:expriments}

\begin{table*}[tb]
    \centering
    \normalsize
    \renewcommand{\arraystretch}{1.3}
    \setlength{\tabcolsep}{3pt} 
    \begin{tabular}{llcccccccc}
        \toprule
        \textbf{Category} & \textbf{Methods}  & \textbf{Paradigm} & \textbf{TOPIQ$\uparrow$}   & \textbf{LPIPS$\downarrow$}   &  \textbf{CMMD$\downarrow$} & \textbf{Mask(\%)$\uparrow$} & \textbf{CLIP.T(\%)$\uparrow$} & \textbf{DSD$\downarrow$} & \textbf{IRS$\uparrow$} \\ 
        \midrule

        \multirow{6}{*}{Comparisons} 
        & TediGAN \shortcite{xia2021tedigan}  &  GAN  &  0.7130   & \underline{0.4885} &  1.562 &   87.89   & 23.90 & \underline{2.46} & -0.1446  \\
        & UaC\shortcite{nair2023unite}  &  Diffusion & 0.6027   & 0.5761  &  1.982  & 84.71 & 25.52  &  3.41  &   -0.4001 \\
        & PixelFace+ \shortcite{du2023pixelface+} &  GAN  & 0.5720    & 0.5640  &  1.273 & \underline{93.82}  & \underline{26.16} & 2.61   &  \underline{0.4608} \\
        & CoDiffusion \shortcite{huang2023collaborative} &  Diffusion &  0.7381    &  0.5845 &  \underline{0.734} & 83.75  &  24.51 & 3.22 &  -0.1265 \\
        & ControlNeXt \shortcite{peng2024controlnext}&  Diffusion & 0.8037    &  0.5469  &  1.042  &  91.65  &  25.88   & 2.66 &  0.3675    \\
        & MM2Latent \shortcite{meng2025mm2latent} & GAN  &  \underline{0.8252}   & 0.5124 &  1.178  &  85.05  &  24.61  &  3.05  &  0.2754 \\ 
        \midrule
      
        \rowcolor{blue!5}
        \textbf{Ours} 
        & \textbf{MDiTFace}  &  Diffusion &  \textbf{0.8466}   & \textbf{0.4725}   &  \textbf{0.482}   &  \textbf{94.64}  &  \textbf{26.75}   & \textbf{2.38}  & \textbf{0.6993}     \\          
     \bottomrule
    \end{tabular}
    \caption{Quantitative comparison with state-of-the-art methods on the MM-CelebA test set. (Bold indicates the best, and underline indicates the second-best.)}
    \label{tab:exp_mm-celeba}
\end{table*}

\subsection{Experimental Setup}
In this section, we provide a comprehensive introduction to the specific experimental settings, encompassing datasets, evaluation metrics, and implementation details.

\textbf{Datasets.} MM-CelebA \cite{lee2020maskgan} is a large-scale, multimodal facial synthesis benchmark dataset widely used for evaluation. It comprises 30,000 highly diverse, high-resolution real-world facial images. Each image is accompanied by 10 distinct textual descriptions and a manually annotated semantic mask. The semantic masks cover 19 semantic categories, including major facial components such as hair, skin, eyes, and nose, as well as accessory categories like glasses and clothing. For model training and testing, the MM-CelebA dataset is split into training and test sets in a 9:1 ratio. In addition, to further evaluate the generalization capability of our model, we extended the FFHQ-Text dataset \cite{zhou2021generative} into a mask-text multimodal version, denoted as MM-FFHQ, where the semantic masks were obtained using a pre-trained facial parser \cite{zheng2022general}.

\textbf{Metrics.} We employ a range of widely recognized metrics to evaluate the performance of our model, spanning image quality, conditional consistency, and human preference. For image quality,we employ the no-reference TOPIQ metric \cite{chen2024topiq} to evaluate the overall quality of synthesized facial images. Additionally, we utilize the CLIP Maximum Mean Discrepancy (CMMD) to quantify the distributional divergence between synthesized and real facial images in the CLIP feature space \cite{jayasumana2024rethinking}, and the Learned Perceptual Image Patch Similarity (LPIPS) to assess perceptual distortion between synthesized and original images \cite{zhang2018unreasonable}. For conditional consistency, we measure semantic alignment between text and images using the CLIP score \cite{radford2021learning}, evaluate pixel-level alignment between segmentation masks and images via Mask Accuracy, and quantify structural similarity between synthesized and real facial images using the DINO Structure Distance (DSD) \cite{tumanyan2022splicing}. Additionally, we employ a pre-trained aesthetic scoring model \cite{xu_imagereward:_2023} to compute the Image Reward Score (IRS), which serves as an indirect proxy for the human-centric visual appeal of the generated images.

\textbf{Implementation Details.} All experiments were conducted on a server equipped with 8 NVIDIA A100 GPUs, each with 80 GB of memory. We employed parameter-efficient LoRA fine-tuning with a rank of 8, and selected Prodigy \cite{mishchenko2024prodigy} as the optimizer. The base learning rate was set to 1.0, and the weight decay coefficient was configured as 0.01. During training, we used a batch size of 16 and optimized the model for 5,000 steps. For inference, we employed the Flow-Matching Euler Discrete scheduler, with a random seed of 42 and 28 sampling steps.
\subsection{Quantitative Evaluation}
Table~\ref{tab:exp_mm-celeba} presents the quantitative comparison on the MM-CelebA benchmark dataset, demonstrating that the proposed MDiTFace outperforms all competing methods across all metrics, thereby establishing a new performance benchmark. Specifically, in terms of image fidelity measured by TOPIQ, LPIPS, and CMMD, MDiTFace achieves improvements of 2.59\%, 3.27\% and 34.33\%, respectively, compared to the second-best method. For conditional consistency evaluated using Mask (\%), CLIP.T (\%), and DSD metrics, MDiTFace shows enhancements of 0.87\%, 2.26\% and 8.81\%, respectively, over the runner-up. This confirms that MDiTFace generates facial images with not only higher fidelity but also better semantic alignment with multimodal input conditions. Additionally, MDiTFace attains the highest Image Reward Score of 0.6993, proving its superior performance in human-centric visual appeal. Table~\ref{tab:exp_mm-ffhq} displays the quantitative evaluation on the MM-FFHQ dataset. MDiTFace continues to maintain leading performance across multiple metrics, demonstrating its robust zero-shot generalization capability. More quantitative experiments are provided in the Supplementary Material.

\begin{table*}[tb]
    \centering
    \normalsize
    \renewcommand{\arraystretch}{1.3}
    \setlength{\tabcolsep}{3pt} 
    \begin{tabular}{llcccccccc} 
        \toprule
        \textbf{Category} & \textbf{Methods} & \textbf{Paradigm}  & \textbf{TOPIQ$\uparrow$}  & \textbf{LPIPS$\downarrow$} & \textbf{CMMD$\downarrow$} & \textbf{Mask(\%)$\uparrow$} & \textbf{CLIP.T(\%)$\uparrow$} & \textbf{DSD$\downarrow$} & \textbf{IRS$\uparrow$} \\ 
        \midrule
        
        \multirow{6}{*}{Comparisons} 
        & TediGAN \shortcite{xia2021tedigan}&  GAN & 0.6992  & 0.6420 & 1.091 & 86.81 & 25.03 & 3.32 & -0.4847 \\
         & UaC \shortcite{nair2023unite}  &  Diffusion & 0.6106 & 0.6752 & 1.796 & 86.17 & 26.97 & 6.38 & -0.8851 \\
        & PixelFace+ \shortcite{du2023pixelface+} &  GAN &  0.5609 & 0.6705 & 1.917 & \underline{95.02} & 26.60 & \underline{3.08} & -0.2616 \\
        & CoDiffusion \shortcite{huang2023collaborative}  &  Diffusion  & 0.5035 & 0.6767 & 2.178 & 81.78 & 23.03 & 3.85 & -1.2101 \\
        & ControlNeXt \shortcite{peng2024controlnext}  &  Diffusion & 0.7594 & \underline{0.6165} & \underline{1.073} & 86.52 & \underline{27.80} & 3.84 & -0.4063 \\
        & MM2Latent \shortcite{meng2025mm2latent}  &  GAN & \underline{0.8198} & 0.6371 & \textbf{0.437} & 84.24 & 26.37 & 3.52 & \underline{-0.1197} \\ 
        \midrule
        
        \rowcolor{blue!5}
        \textbf{Ours} 
        & \textbf{MDiTFace}   &  Diffusion & \textbf{0.8503} & \textbf{0.4969} & 1.135 & \textbf{96.13} & \textbf{28.05} & \textbf{2.79} & \textbf{0.0842} \\
    
        \bottomrule
    \end{tabular}
    \caption{Quantitative comparison with state-of-the-art methods on the MM-FFHQ  dataset.}
    \label{tab:exp_mm-ffhq}
\end{table*}

\subsection{Qualitative Comparison}
Figure \ref{fig:comparision_with_sota_methods} displays the facial images synthesized by the proposed MDiTFace and competing methods under the same multimodal conditions. It is evident that the facial images synthesized by MDiTFace exhibit higher fidelity. Meanwhile, in terms of fine-grained attributes such as earrings, hair color, and hat shape, MDiTFace can better align with the input conditions, as shown in rows 1-3. This is consistent with the results of quantitative evaluation experiments. In contrast, existing methods suffer from various deficiencies. For instance, methods like UaC \cite{nair2023unite}, PixelFace+ \cite{du2023pixelface+}, and CoDiffusion \cite{huang2023collaborative} generate facial images with noticeably inadequate fidelity, resulting in unsatisfactory image quality. Other methods, such as TediGAN \cite{xia2021tedigan}, ControlNeXt \cite{peng2024controlnext}, and MM2Latent \cite{meng2025mm2latent}, encounter challenging issues of attribute loss or drift, leading to discrepancies between the synthesized facial images and the expected input conditions. Additionally, when handling the task of facial synthesis under extreme poses, as depicted in row 4, MDiTFace still maintains significant robustness compared to existing methods. More examples can be found in the Supplementary Material.

\begin{figure}[h]
\centering
\includegraphics[width=0.95\linewidth]{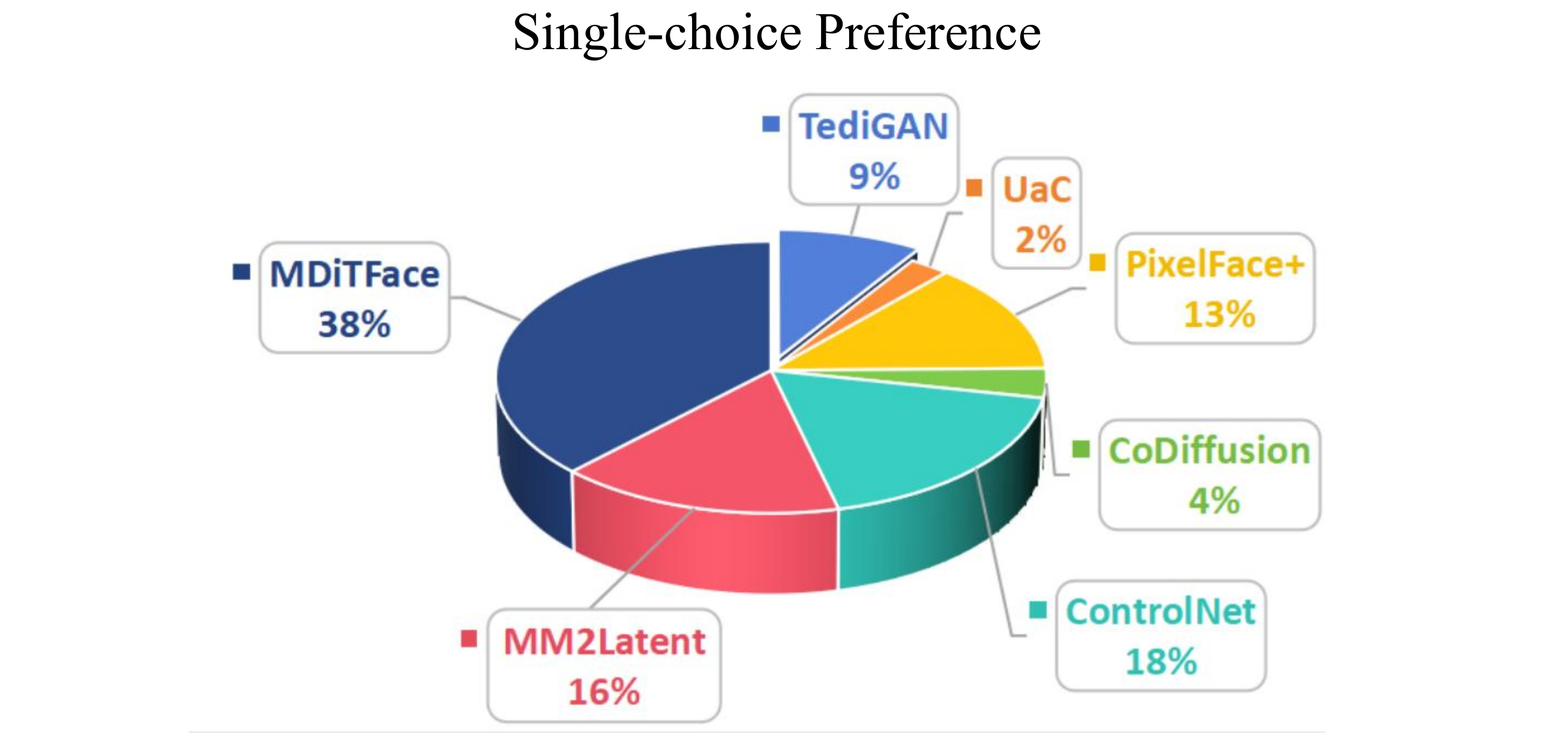}
\caption{User study.
Our method secured the highest proportion of user support.}
\label{fig:user-study}
\end{figure}
\subsection{User Study} We conducted a user study with 30 participants to estimate choice preferences from a human perspective. We randomly selected 50 sets of test data and shuffled the order of facial images synthesized by all the methods. Participants were required to choose one facial image that best met the multi-constraint conditions of masks and text prompts. Figure \ref{fig:user-study} presents the statistical results. Our method received the highest proportion (38\%) of user support, far outperforming the second-place method (ControlNeXt, 18\%) and the third-place method (MM2Latent, 16\%). These findings further validate the effectiveness of MDiTFace in mask-text collaborative facial synthesis.
\subsection{Ablation Studies}
We conducted ablation studies to comparatively analyze different mask-handling approaches, validate the effectiveness of our decoupled attention mechanism, and assess the impact of hyperparameter settings in our experiments.
\begin{table}[tb]
\centering
    \setlength{\tabcolsep}{5pt} 
    \normalsize
    \renewcommand{\arraystretch}{1.3}
    \begin{tabular}{ccccc} 
    \toprule
     Process  & Mask(\%)$\uparrow$ & CLIP.T(\%)$\uparrow$ & IRS$\uparrow$ &CMMD$\downarrow$  \\ \midrule
     Concat & 88.73  &  26.08  &  0.3699  &  \textbf{0.476}  \\
     \rowcolor{blue!5}
     Ours   &  \textbf{94.64} &  \textbf{26.75}  & \textbf{0.6993}  &  0. 482\\ \bottomrule
    \end{tabular}
\caption{Effectiveness comparison of different mask condition processing approaches.}
\label{tab:mask_condition_processing}
\end{table}

\textbf{Handling of the mask condition.}
Table~\ref{tab:mask_condition_processing} presents a quantitative comparison of different mask condition processing approaches. The method denoted by \textit{Concat} is as described in Equation \ref{eq:mask_concat}, which simply performs channel-wise concatenation of the noisy image latent and the mask. The experimental results demonstrate that this processing approach leads to suboptimal performance in terms of mask condition consistency, achieving a score of only 88.73 on the Mask(\%) metric. This represents a 5.91\% gap compared to the unified tokenization method we employ. These findings provide compelling evidence for the superiority of using unified tokenization to handle multimodal conditions in mask-text collaborative face generation tasks.

\begin{figure}[t]
\centering
\includegraphics[width=0.95\linewidth]{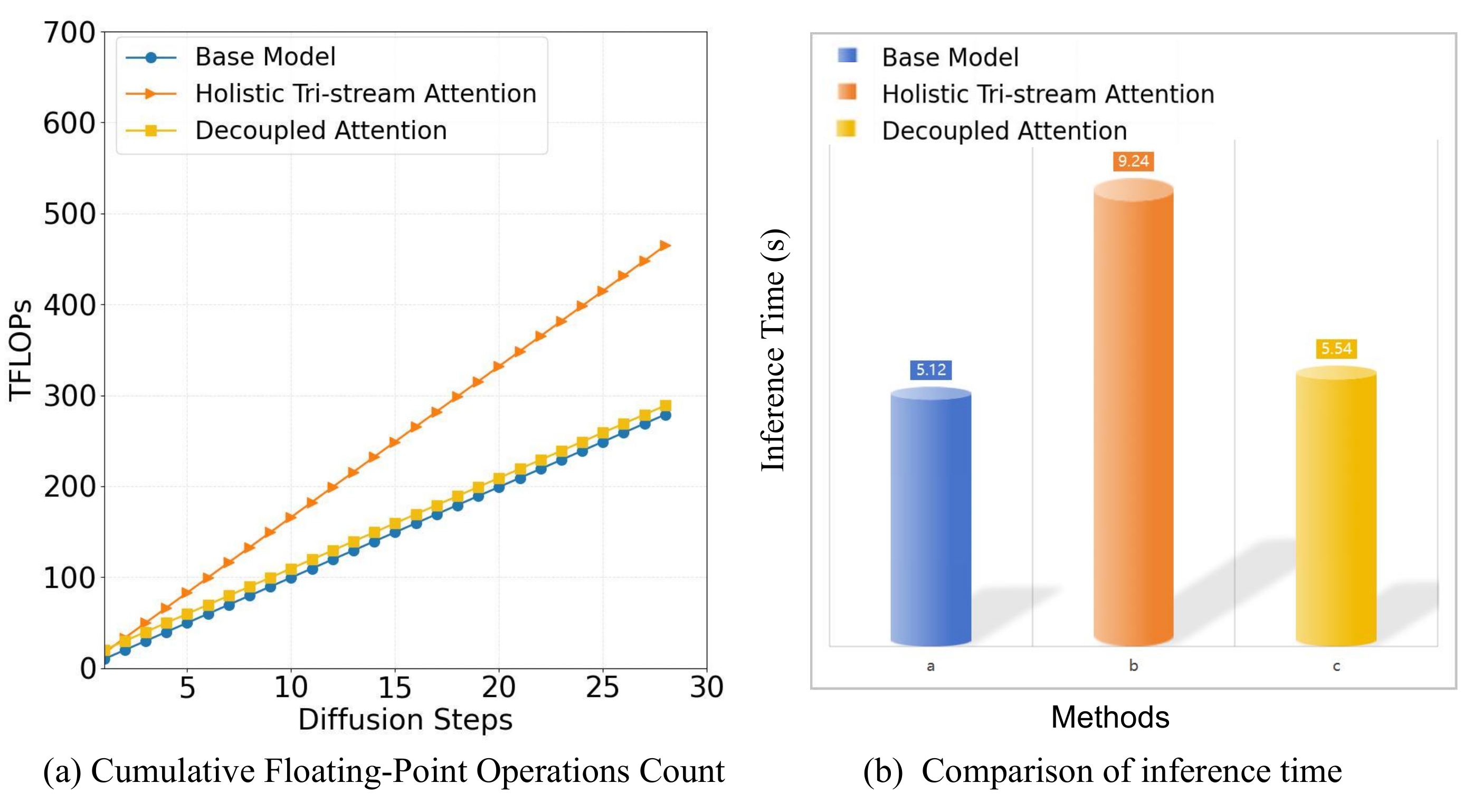}
\caption{Additional computational overhead introduced by mask condition.}
\label{fig:efficiency_comparison}
\end{figure}
\textbf{Effectiveness of Decoupled Attention.}
Table \ref{tab:effectiveness_of_decoupled_attention} presents a systematic quantitative comparison across the three multivariate transformer architectures depicted in Figure \ref{fig:mm-attention-evolutions}(b)-(d). The holistic tri-stream attention mechanism (Figure \ref{fig:mm-attention-evolutions}(b)) incurs an additional computational overhead of 185.79 TFLOPs due to its simultaneous processing of mask conditions. The hard-decoupled variant (Figure \ref{fig:mm-attention-evolutions}(c)) effectively reduces computational load by eliminating implicit coupling between static pathways and temporal embeddings, but at the expense of weakened mask-text interaction capabilities, as evidenced by a 3.5\% absolute decline in the Mask(\%) metric (from 94.72 to 91.22). In contrast, our improved decoupled attention mechanism (Figure \ref{fig:mm-attention-evolutions}(d)) reconstructs perceptual pathways from mask tokens to text tokens by introducing joint multimodal condition tokens ⟨CM, CT⟩ into static pathways. This approach maintains statistical equivalence in model performance while dramatically reducing mask-induced computational overhead from 185.79 TFLOPs to 9.95 TFLOPs (a 94.7\% reduction). Figure \ref{fig:efficiency_comparison} further validates the significant computational efficiency gains through cumulative floating-point operation statistics and inference time comparisons.
\begin{table}[!t]
\centering
    \normalsize
    \renewcommand{\arraystretch}{1.3}
    \setlength{\tabcolsep}{1.5pt} 
    \begin{tabular}{cccccc} 
    \toprule
        Type & Mask(\%)$\uparrow$ & CLIP.T(\%)$\uparrow$ & IRS$\uparrow$ &CMMD$\downarrow$ & $\Delta$TFLOPs \\ \midrule
      (b) & 94.72  & 26.86 &  0.6967 &  0.524 & \underline{185.79} \\ \midrule
      (c) & \underline{91.22} & 26.80  & 0.7039  &  0.602 & 6.64  \\
      \rowcolor{blue!5}
      (d)  & 94.64  & 26.75 & 0.6993  & 0.482 & 9.95 \\ \bottomrule
    \end{tabular}

\caption{Comprehensive quantitative analysis on the effectiveness of the proposed decoupled attention mechanism.}
\label{tab:effectiveness_of_decoupled_attention}
\end{table}
\begin{table}[!t]
\centering
    \normalsize
    \renewcommand{\arraystretch}{1.3}
    \setlength{\tabcolsep}{5pt} 
    \begin{tabular}{cccc} 
    \toprule
        $r$ & Mask(\%)$\uparrow$ & CLIP.T(\%)$\uparrow$ &CMMD$\downarrow$ \\ \midrule
      4 & 94.78  & 26.52 & 0.477  \\ 
      \rowcolor{blue!5}
      8 & 94.64 & 26.75 &  0.482  \\
      16  & 94.64  & 26.70 & 0.470 \\ 
      32  & 94.73  & 26.63 & 0.494 \\ \bottomrule
    \end{tabular}
\caption{Ablation study on LoRA rank setting.}
\label{tab:ablation_of_lora}
\end{table}

\begin{table}[!t]
\centering
    \normalsize
    \renewcommand{\arraystretch}{1.3}
    \setlength{\tabcolsep}{5pt} 
    \begin{tabular}{cccc} 
    \toprule
        $p$ & Mask(\%)$\uparrow$ & CLIP.T(\%)$\uparrow$ &DSD$\downarrow$ \\ \midrule
      0.1 & 94.64 & 26.75 & 2.38  \\ 
      0.3 & 94.21 & 26.68 & 2.40   \\
      0.5  & 94.21  & 26.71 & 2.41 \\ 
      \rowcolor{blue!5}
      0.7  & 93.61  & 26.90 & 2.54 \\ 
      \rowcolor{blue!5}
      0.9  & 84.00  & 27.02 & 3.11 \\\bottomrule
    \end{tabular}
\caption{Ablation study on hyperparameter $p$ of the stochastic condition dropout probability.}
\label{tab:ablation_of_p}
\end{table}
\textbf{The setting of hyperparameters $r$ and $p$.}
We conducted a comparative analysis of model performance under different settings of the LoRA rank ($r$), with the relevant results presented in Table \ref{tab:ablation_of_lora}. The experiments revealed that increasing the value of $r$ did not significantly enhance model performance, which, to a certain extent, confirms the robustness of our proposed method. All quantitative scores reported in this paper were obtained when the LoRA rank was set to $8$. Additionally, we introduced a stochastic condition dropout strategy during the training phase, aiming to enable the model to simultaneously adapt to facial generation tasks driven by unimodal conditions. We also conducted a comparative analysis of the impact of the hyperparameter $p$ setting on model performance, with the relevant experimental results shown in Table \ref{tab:ablation_of_p}. It can be observed that, under a fixed number of fine-tuning steps, as the value of the hyperparameter $p$ increases, the model's learning efficiency for newly introduced mask conditions significantly decreases. This trend is particularly evident in the Mask(\%) and DSD metrics when $p$ exceeds $0.5$.

\subsection{Multimodal compatibility}
\begin{figure}[h]
\centering
\includegraphics[width=0.95\linewidth]{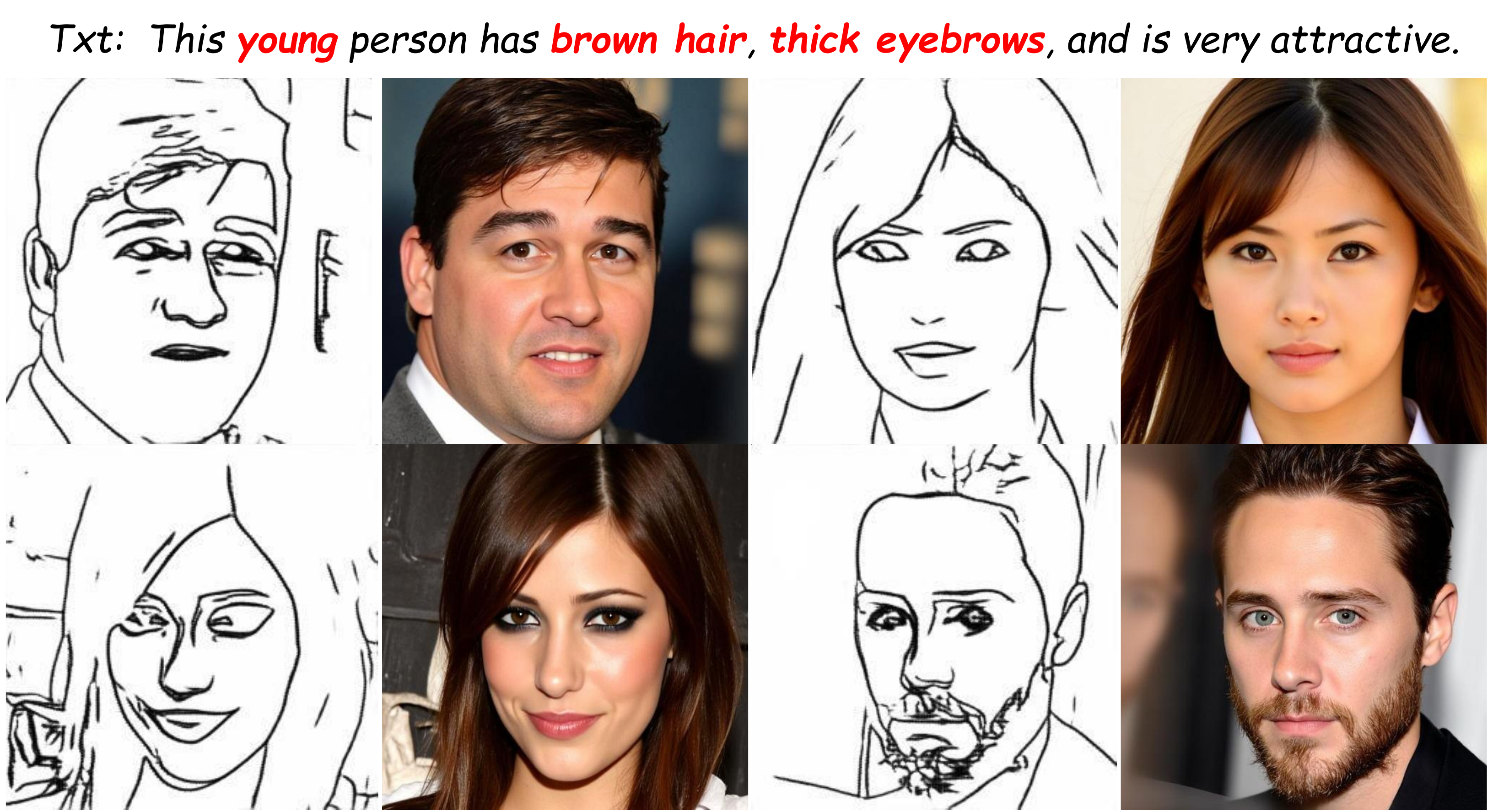}
\caption{Sketch-text jointly driven facial synthesis.}
\label{fig:sketch2face}
\end{figure}

Although the proposed MDiTFace method is primarily tailored for the mask-text collaborative facial generation task, its intrinsic architecture demonstrates robust generalization capabilities, allowing it to seamlessly accommodate combined inputs from diverse multimodal conditions. For example, multimodal pairings such as depth-text and sketch-text can all be efficiently processed by MDiTFace. As depicted in Figure \ref{fig:sketch2face}, MDiTFace exhibits exceptional performance in the sketch-text collaborative facial generation task. In this scenario, the sketch provides coarse contour and structural information of the face, while the text serves to refine facial attributes, including skin tone, expressions, and fine-grained feature details. MDiTFace precisely captures key information from both the sketch and text, organically fusing them to generate high-quality, highly realistic facial images that closely align with the input descriptions. This fully underscores its powerful compatibility in handling multimodal condition combinations.

\section{Conclusion}
This study proposes MDiTFace, a high-fidelity mask-text collaborative facial synthesis framework built upon a customized diffusion transformer. The framework employs unified tokenization to process multimodal conditions, effectively bridging the representational gaps between heterogeneous modalities. The restructured multivariate transformer blocks is capable of efficiently and synchronously handling both mask and text tokens. Its internally designed decoupled attention mechanism not only facilitates flexible interactions among multimodal features but also explicitly partitions the computational flow into dynamic and static pathways by decoupling implicit dependencies between mask tokens and temporal embeddings. Relevant features in the static pathway only need to be computed once and then cached for reuse across denoising steps. This mechanism reduces redundant computational overhead introduced by mask conditions by over 94\% while maintaining model performance, thereby significantly enhancing the efficiency of facial synthesis.

\FloatBarrier

\bibliography{aaai2026}
\newpage

\newpage
\appendix

\section{Supplementary Material}
This supplementary material offers a thorough, multi-dimensional analysis of the proposed MDiTFace method, emphasizing its key features and strengths. To verify its generalization capacity, we conduct additional quantitative experiments and provide more visual comparisons with cutting-edge approaches, demonstrating its high fidelity in facial image generation and strict compliance with given constraints. We also conduct experiments to evaluate MDiTFace's performance under multi-modal constraints, with a particular focus on its robustness and diversity in handling complex inputs. For unimodal-driven (text or mask) scenarios, both quantitative and qualitative results confirm its outstanding generalization ability in facial image synthesis. Finally, we discuss the limitations of the proposed MDiTFace method and propose directions for future research.
\begin{table*}[!h]
    \centering
    \normalsize
    \renewcommand{\arraystretch}{1.3}
    \setlength{\tabcolsep}{3pt} 
    \begin{tabular}{llcccccccc} 
        \toprule
        \textbf{Category} & \textbf{Methods} & \textbf{Paradigm}  & \textbf{TOPIQ$\uparrow$} & \textbf{LPIPS$\downarrow$} & \textbf{CMMD$\downarrow$} & \textbf{Mask(\%)$\uparrow$} & \textbf{CLIP.T(\%)$\uparrow$} & \textbf{DSD$\downarrow$} & \textbf{IRS$\uparrow$} \\ 
        \midrule
        
        \multirow{6}{*}{Comparisons} 
        & TediGAN \shortcite{xia2021tedigan}  &  GAN & 0.6707  & 0.7257 & 2.367 & 71.55 & 22.44 & 5.50 & -0.2374 \\
        & UaC \shortcite{nair2023unite}   &  Diffusion & 0.3647  & \underline{0.6983} & 3.636 & 74.33 & 22.38 & 4.66 & -1.2595 \\
        & PixelFace+ \shortcite{du2023pixelface+}  &  GAN & 0.3214   & 0.6987 & 3.304 & \underline{91.20} & 23.91 & \underline{4.65} & -1.2848 \\
        & CoDiffusion \shortcite{huang2023collaborative}  &  Diffusion  & 0.3825   & 0.7556 & 2.221 & 53.44 & 21.83 & 5.85 & 1.4839 \\
        & ControlNeXt \shortcite{peng2024controlnext}  &  Diffusion & 0.5483 & 0.7576 & \underline{1.138} & 74.14 & \textbf{27.05} & 5.12 & -0.5563 \\
        & MM2Latent \shortcite{meng2025mm2latent} &  GAN & \textbf{0.7552}  & 0.7179 & 1.390 & 67.43 & 24.05 & 5.21 &  \underline{-0.0827} \\ 
        \midrule
        
        \rowcolor{blue!5}
        \textbf{Ours} 
        & \textbf{MDiTFace}   &  Diffusion & \underline{0.7182} & \textbf{0.6157} & \textbf{1.033} & \textbf{95.94} & \underline{26.40} & \textbf{4.22} & \textbf{0.2424} \\
    
        \bottomrule
    \end{tabular}
    \caption{Quantitative comparison with state-of-the-art methods on the MM-FairFace  dataset.}
    \label{tab:exp_mm-fairface}
\end{table*}

\subsection{Additional Quantitative Experiments}
Table \ref{tab:exp_mm-fairface} presents the quantitative evaluation results of our proposed MDiTFace method for multimodal facial synthesis, with experiments carried out on the MM-FairFace dataset. Built on the FairFace dataset \cite{karkkainen2021fairface} (released by MIT in 2019 to reduce race-, gender-, and age-related biases in face recognition), MM-FairFace inherits race, gender, and age annotations for each facial image. We constructed MM-FairFace by using a pre-trained facial parser \cite{zheng2022general} to generate semantic masks for each image, following a methodology similar to that for the MM-FFHQ dataset.

The experimental results clearly show that MDiTFace outperforms most competing methods across the majority of evaluation metrics, demonstrating its strong generalization ability. For instance, on the Mask(\%) metric, it achieves a score of 95.95, a 4.74\% improvement over the second-best method. Similarly, on the LPIPS and CMMD metrics, it surpasses the second-best method by 11.83\% and 9.23\%, respectively. These results indicate that MDiTFace not only better understands and aligns multimodal input conditions but also generates facial images with higher fidelity, closely resembling real facial images.

\subsection{Qualitative Comparisons with SOTA Methods}
\begin{figure*}[h]
\centering
\includegraphics[width=0.98\textwidth]{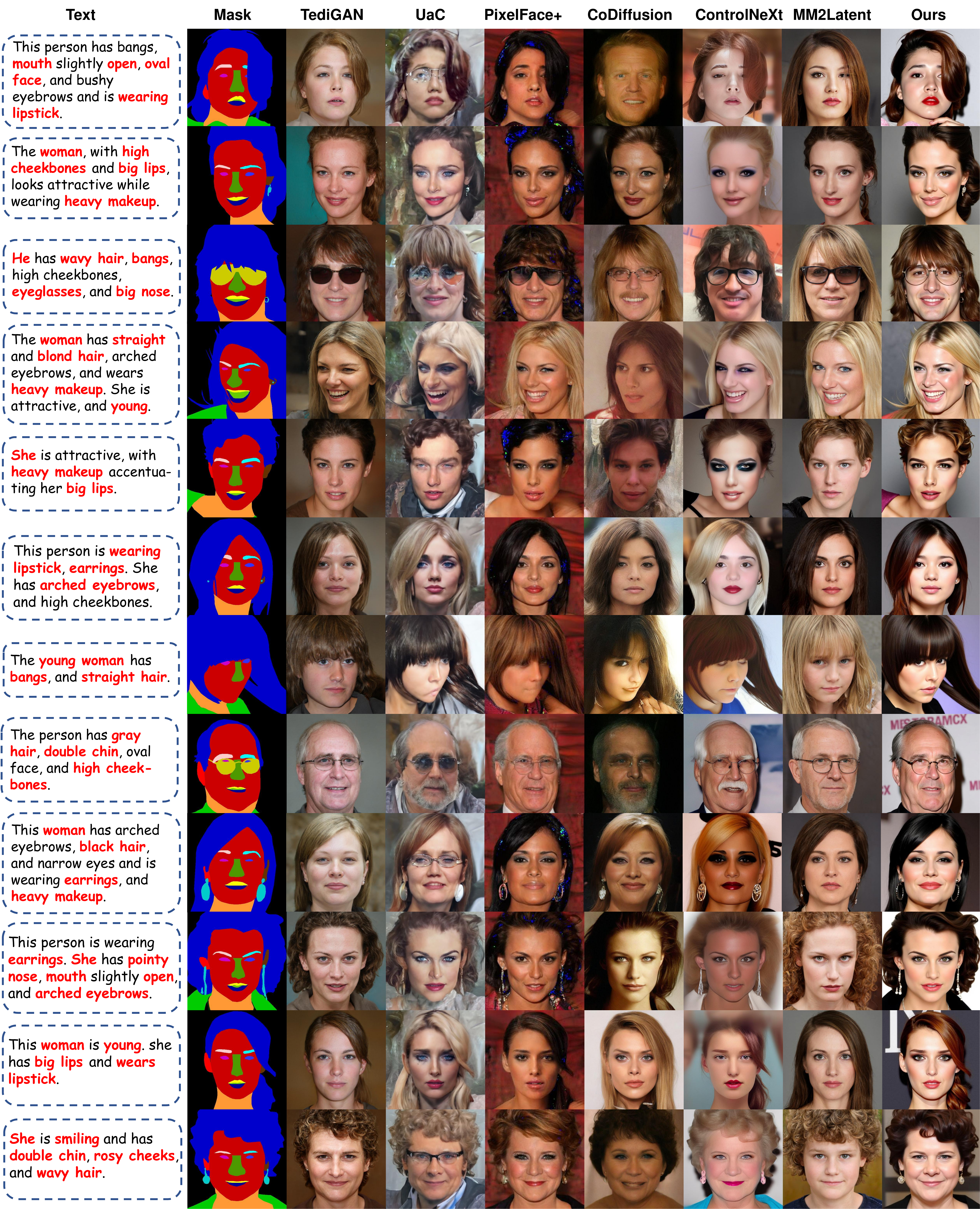} 
\caption{Qualitative comparison with state-of-the-art methods on the MM-CelebA dataset.}
\label{fig:visual_comparison_on_MM-CelebA}
\end{figure*}

\begin{figure*}[h]
\centering
\includegraphics[width=0.98\textwidth]{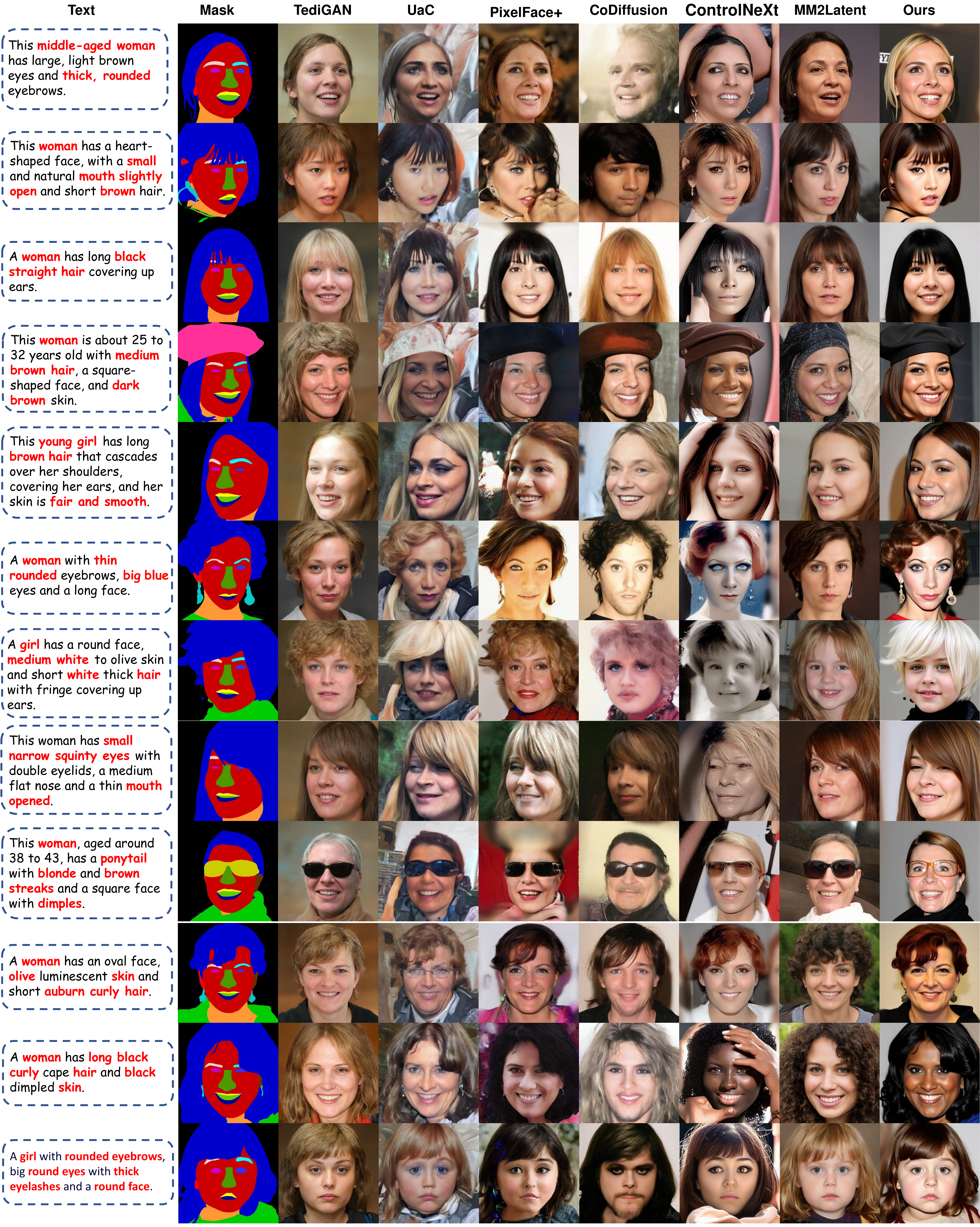} 
\caption{Qualitative comparison with state-of-the-art methods on the MM-FFHQ dataset.}
\label{fig:visual_comparison_on_MM-FFHQ}
\end{figure*}

\begin{figure*}[h]
\centering
\includegraphics[width=0.98\textwidth]{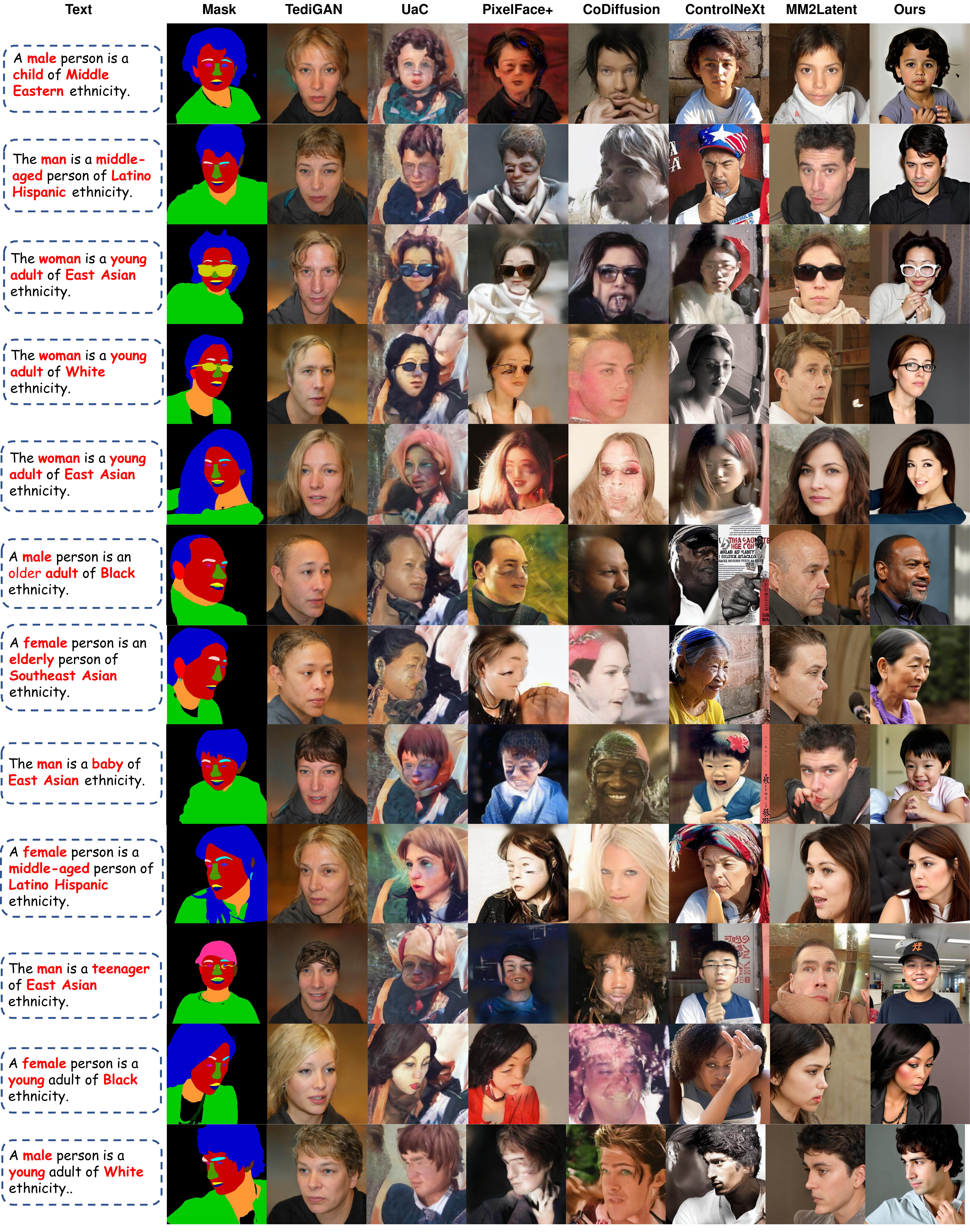} 
\caption{Qualitative comparison with state-of-the-art methods on the MM-FairFace dataset.}
\label{fig:visual_comparison_on_MM-FairFace}
\end{figure*}

Figures \ref{fig:visual_comparison_on_MM-CelebA}, \ref{fig:visual_comparison_on_MM-FFHQ}, and \ref{fig:visual_comparison_on_MM-FairFace} display visual comparisons of typical facial images synthesized by our proposed MDiTFace method and competing state-of-the-art approaches on the MM-CelebA \cite{lee2020maskgan}, MM-FFHQ \cite{zhou2021generative}, and MM-FairFace \cite{karkkainen2021fairface} datasets, respectively. Visually, our method yields facial images with superior naturalness and realism, accompanied by a marked reduction in artifacts. Moreover, compared to other methods, MDiTFace shows distinct advantages in semantic mask alignment (reflecting spatial layout) and fine-grained attribute matching (e.g., hair color, accessories, facial makeup).

Notably, on the MM-FairFace dataset \cite{karkkainen2021fairface}, many competing methods exhibit facial synthesis collapse due to their failure to interpret racial annotations, resulting in poor generalization. In contrast, MDiTFace shows superior robustness by rationally deducing skin tone from textual racial cues, highlighting its strong generalization in textual semantic comprehension.

\subsection{Robustness under Multimodal Combination}
\begin{figure*}[h]
\centering
\includegraphics[width=0.9\textwidth]{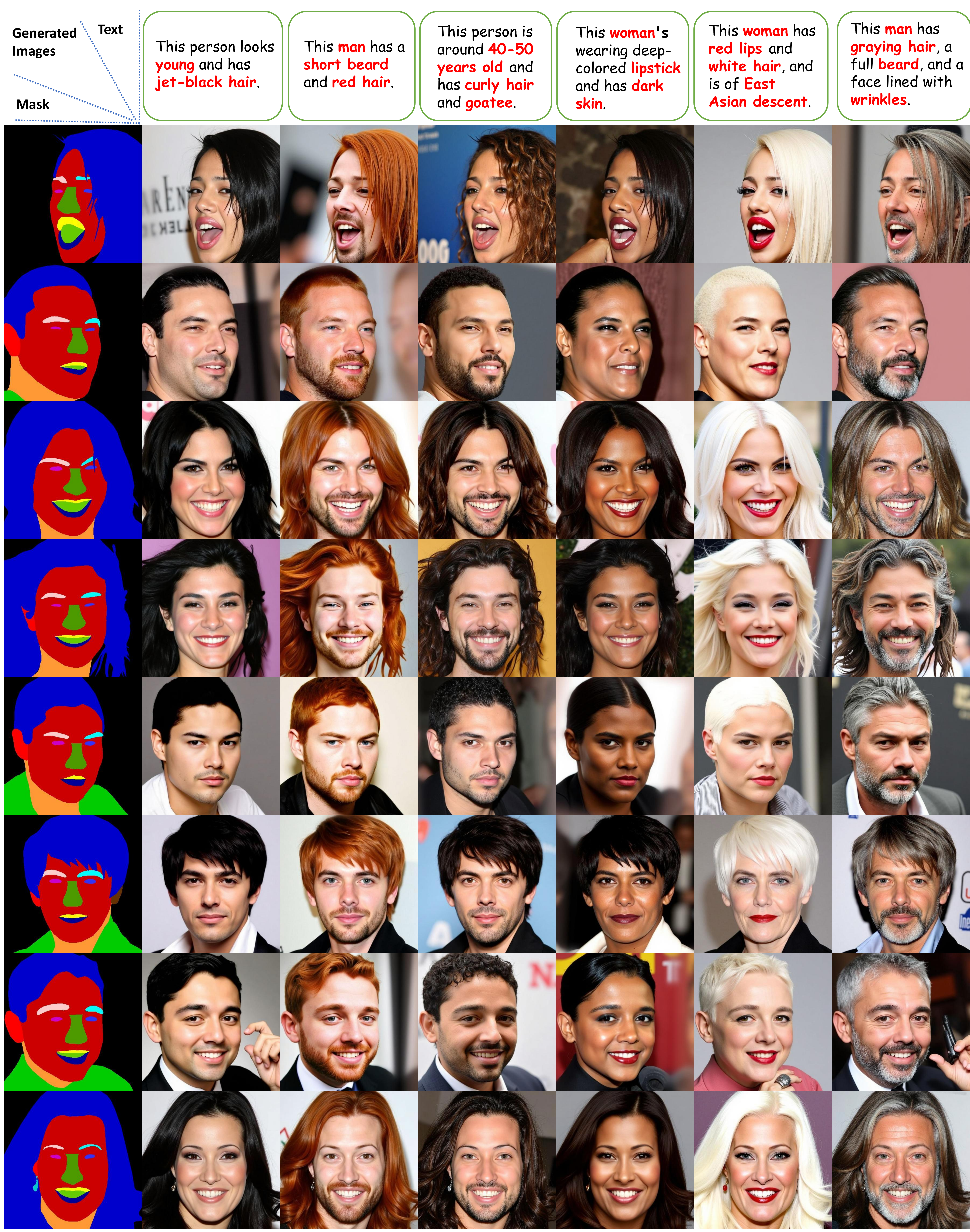} 
\caption{Robustness validation of MDiTFace: collaborative facial synthesis examples with arbitrary mask-text combinations.}
\label{fig:robust_test_of_arbitrary_combinations}
\end{figure*}

To validate the robust performance of the proposed MDiTFace method in facial synthesis under multi-modal condition constraints, we randomly selected $m$ mask conditions and $n$ text conditions, and constructed $m \times n$ condition combinations. Figure \ref{fig:robust_test_of_arbitrary_combinations} displays the facial images generated based on these combinations. When observed row by row, under the same mask condition, as the text conditions vary in aspects such as age, gender, and makeup, MDiTFace can well maintain the alignment characteristics of these semantics. Even in unconventional cases, such as women with extremely short hair or men with extremely long hair, the synthesized facial images remain natural and realistic. When observed column by column, under the same text condition, the pose and contour of the synthesized face change with the variation of mask conditions, which fully demonstrates the method's strong spatial position alignment capability. The above-mentioned observations indicate that MDiTFace exhibits remarkable robustness under condition combinations.

\subsection{Diversity in Facial Generation}
\begin{figure*}[h]
\centering
\includegraphics[width=0.9\textwidth]{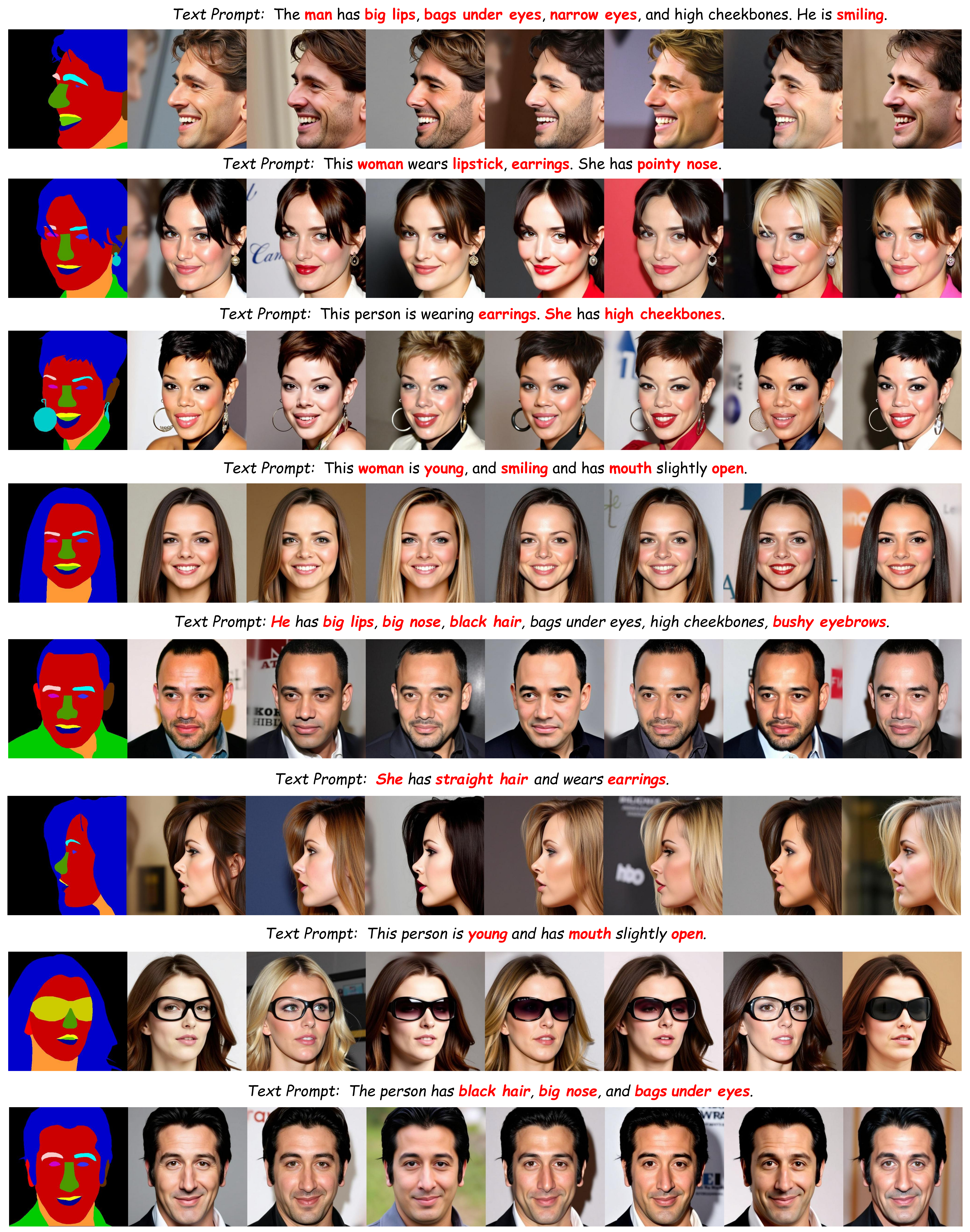} 
\caption{Investigating the generative diversity of MDiTFace under mask-text collaborative constraints.}
\label{fig:generative_diversity}
\end{figure*}

As shown in Fig. \ref{fig:generative_diversity}, we conduct an in-depth analysis of the facial image generation diversity of the proposed MDiTFace under the collaborative mask-text constraints. The experimental results clearly demonstrate that, while adhering to the spatial layout constraints set by the mask conditions and the semantic attribute constraints specified by the text conditions, our MDiTFace can still maintain a high level of generation diversity across other attribute dimensions of facial images. Specifically, these variable attributes cover multiple aspects, such as variations in facial skin tone and hair color, diverse styles of accessories like earrings and glasses, and flexible background transformations. This notable characteristic holds immense application potential in practical scenarios such as data augmentation and interactive creation.
\begin{table*}[!h]
    \centering
    \normalsize
    \renewcommand{\arraystretch}{1.2}
    \setlength{\tabcolsep}{6pt} 
    \begin{tabular}{lcccc} 
        \toprule
         \textbf{Methods} & \textbf{TOPIQ$\uparrow$} & \textbf{CMMD$\downarrow$} & \textbf{CLIP.T(\%)$\uparrow$} & \textbf{IRS$\uparrow$} \\ 
        \midrule
        Clip2Latent \shortcite{pinkneyclip2latent} & \underline{0.7861}   & \underline{1.410} & \underline{27.56} & \textbf{1.0129} \\
        GCDP \shortcite{park2023learning}& 0.6118  & 1.456 & 25.94 & 0.4563 \\
        E3-FaceNet \shortcite{zhang2024fast} & 0.7319  & 2.749  & \textbf{27.94} & 0.8150 \\
        \midrule  
        \rowcolor{blue!5}
        \textbf{Ours} 
        \textbf{MDiTFace} & \textbf{0.8086} & \textbf{0.609} & 27.25 & \underline{0.8569} \\
        \bottomrule
    \end{tabular}
    \caption{Performance comparison with text-only facial generation methods.}
    \label{tab:exp_text-only-driven}
\end{table*}
\begin{table*}[!h]
    \centering
    \normalsize
    \renewcommand{\arraystretch}{1.2}
    \setlength{\tabcolsep}{6pt} 
    \begin{tabular}{lcccc} 
        \toprule
         \textbf{Methods} & \textbf{TOPIQ$\uparrow$} & \textbf{CMMD$\downarrow$} & \textbf{Mask(\%)$\uparrow$} & \textbf{DSD$\downarrow$}  \\
        \midrule
        INADE \shortcite{tan2021diverse} & 0.6156 & 1.871 & 93.60 & 2.57 \\     E2Style \shortcite{wei2022e2style}   & \underline{0.8007} & \underline{1.129} & 90.68 & \underline{2.36} \\
        SemFlow+ \shortcite{wang2024semflow} & 0.7098  & 1.767 & \underline{94.35} & \textbf{2.30}\\
        \midrule  
        \rowcolor{blue!5}
        \textbf{Ours} 
        \textbf{MDiTFace} & \textbf{0.8648} & \textbf{0.478} & \textbf{94.74} & 2.39\\
        \bottomrule
    \end{tabular}
    \caption{Performance comparison with mask-only facial generation methods.}
    \label{tab:exp_mask-only-driven}
\end{table*}
\subsection{Support for Unimodal Driving}
During training, we employed a stochastic condition dropout strategy, enabling MDiTFace to theoretically learn facial synthesis using either text or mask conditions alone. This hypothesis was rigorously validated through both quantitative and qualitative experiments.

\begin{figure*}[h]
\centering
\includegraphics[width=0.9\textwidth]{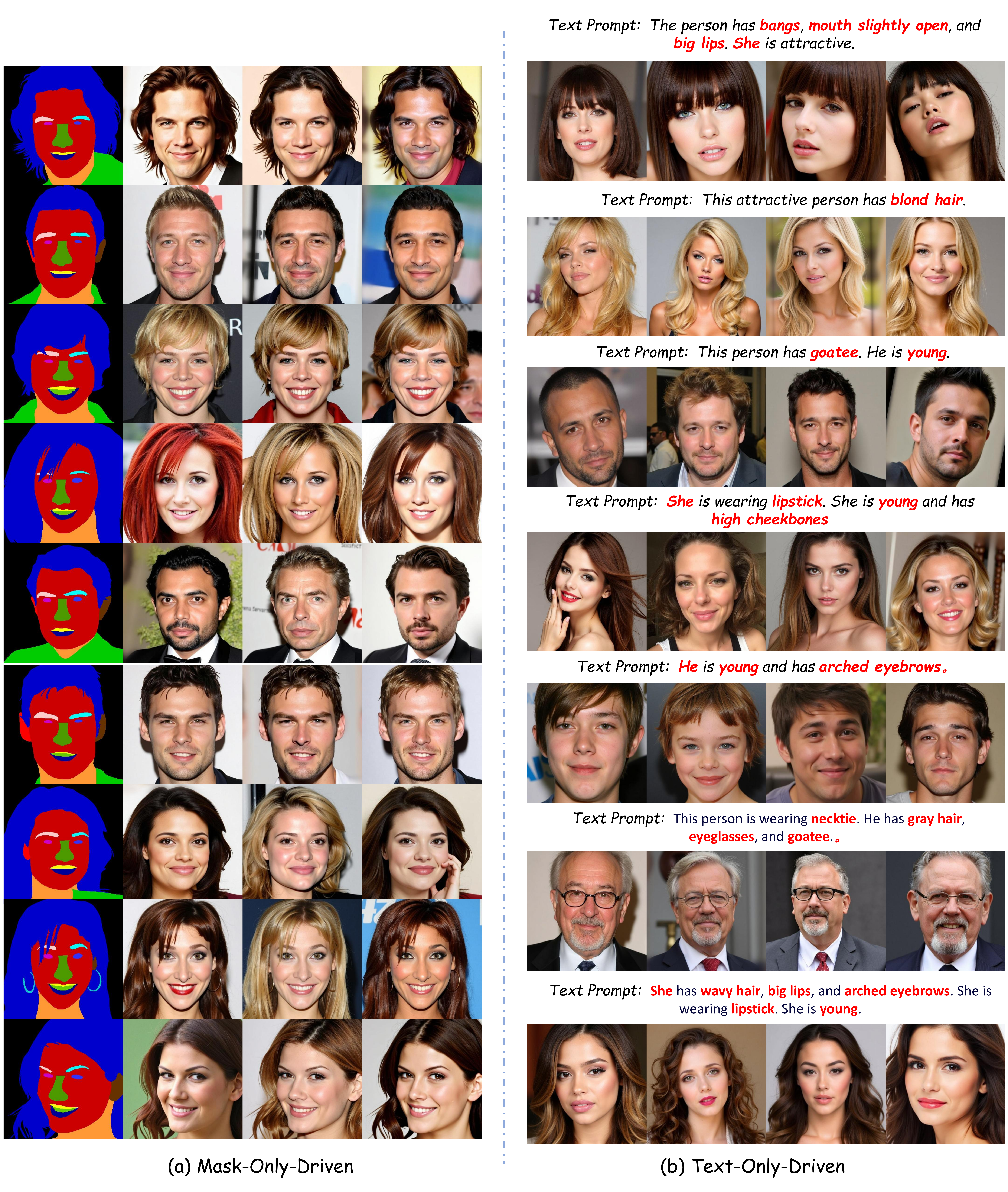} 
\caption{Examples of unimodal-driven facial synthesis supported by MDiTFace.}
\label{fig:unimodal-driven}
\end{figure*}

Quantitatively, as depicted in Tables \ref{tab:exp_text-only-driven} and \ref{tab:exp_mask-only-driven}, we conducted a comparison between MDiTFace and state-of-the-art specialized facial generation models under text-only and mask-only conditions. Although these competing models were specifically designed for particular modalities, MDiTFace achieved comparable performance and outperformed them significantly in terms of TIPIQ, CMMD, and Mask(\%) metrics. This highlights its remarkable generalization capability in unimodal-driven synthesis. Qualitatively, Figure \ref{fig:unimodal-driven} showcases facial images generated exclusively from masks or text descriptions. These visual results further demonstrate MDiTFace's proficiency in achieving diverse facial synthesis under single-condition constraints.

\subsection{Limitations and Prospects}
Although this study has made significant strides in the field of multimodal collaborative facial synthesis by utilizing mask-text inputs, there are still several promising directions for further improvement.

\textbf{Modality Combination Exploration.} Currently, our experimental validation is confined to the fusion of mask and text modalities. Although the proposed MDiTFace architecture is inherently general and theoretically amenable to diverse modality combinations, we have not yet extended our investigations to incorporate other potential pairings, such as sketch-text and keypoint-text. Expanding the scope of modality combinations could unlock new capabilities and applications for our model.

\textbf{Computational Efficiency Optimization.} To improve computational efficiency, we have introduced a structured decoupled attention mechanism. This mechanism segregates internal computations into dynamic and static pathways, contingent on whether timestep embeddings modulate the process. By caching computational features within the static pathway and reusing them across denoising steps, we have effectively reduced redundant computations related to the mask condition by over 94\%. Nevertheless, the iterative denoising nature inherent to diffusion models results in MDiTFace's inference time cost remaining substantially higher than that of comparable generative adversarial network-based methods. To mitigate this limitation, future work could explore the integration of techniques such as accelerated sampling or sparse attention to further enhance the generation efficiency of MDiTFace and alleviate the current constraints on its inference speed.


\end{document}